# Competency Tracking for English as a Second or Foreign Language Learners

Robert Bishop
rbishop@yokohama-cu.ac.jp
robertleebishop@gmail.com

Abstract

My system utilizes the outcomes feature found in Moodle and other learning content management systems (LCMSs) to keep track of where students are in terms of what language competencies they have mastered and the competencies they need to get where they want to go. These competencies are based on the Common European Framework for (English) Language Learning. This data can be available for everyone involved with a given student's progress (e.g. educators, parents, supervisors and the students themselves). A given student's record of past accomplishments can also be meshed with those of his classmates. Not only are a student's competencies easily seen and tracked, educators can view competencies of a group of students that were achieved prior to enrollment in the class. This should make curriculum decision making easier and more efficient for educators.

Introduction

When you are learning a second language there are a few facts that, if they are made easily seen and shared, can make a big difference in your efforts to reach your goals. It's important to know; where you've been, where you are, where you want to go, and how you are going to get there. My system utilizes features found in Moodle and other learning content management systems (LCMSs) that keep track of where students are in terms of what language competencies they have mastered and the competencies they need to get where they want to go. This data can be available for everyone involved with a given student's progress (e.g. educators, parents, supervisors and the students themselves). A given student's record of past accomplishments can also be meshed with those of his classmates. Not only are a student's competencies easily seen and tracked, educators can view competencies of a group of students that were achieved prior to enrollment in the class. This should make curriculum decision making easier and more efficient for educators.

Although no one follows a completely predictable course in their language development, language learners tend to acquire skills in a similar order (Dulay & Burt). In addition to this phenomenon, there are certain aspects of language that should take priority due to the frequency of their usage. For example, it has been my experience that students learn how to use simple past long before embedded questions. A curriculum that incorporates these two issues would prove most efficient.

This concept is nothing new. Most educators and textbook writers make assumptions about what students know at what level and how that compares with the skills they should have. What educators are often still in the dark about, however, is precisely what skills were learned outside their respective classes. Unless they have been following students through the same program (with excellent record keeping), they can only guess when a skill was learned and how well a student mastered it. As an TESL instructor myself, I often wonder if I am repeating old material. Students themselves may know that they are repeating curriculum but may wonder what they *should* be studying next.

Using a Learning Content Management System for Competency Tracking

Due to ongoing developments in speed and ubiquity of the internet combined with powerful LCMSs such as Moodle or Blackboard, it is now possible to keep detailed, sharable records of what a student has achieved. Many programs have developed their own databases for this purpose but they have for the most past remained in-house. Since the system I am proposing can be used online, what a student has learned can be easily tracked, updated and shared. My system is also standardized so it can be set up easily in any language program. Not only can a student's progress be tracked within a given school, a student can save and move his data from one program to another. If a student uses my proposed system in one school, his data can be uploaded to another school's LCMS. In terms of classes, a teacher could see where a group of his newly enrolled students stand language wise at a glance.

Competencies and the Common European Framework for Language Learning

Before going further, I should make clear what I mean by competencies and define them. A 'competency is a set of defined behaviors that provide a structured guide enabling identification, evaluation and  development of the behaviors in [an] individual[sic]' (Raven & Stephenson 2001). In the context of learning a language, these behaviors would include, for example, using different grammar points, performing linguistic functions, being able to use different lexical items and a number of other behaviors involved in using language. So far, the competencies in my system include grammar points and functions. My system tracks these competencies from beginner level to advanced.

The competencies I have included in my system all come from the Common European Framework (CEFR). For those who are unaware of the CEFR,  '[It] is a guideline used to describe achievements of learners of foreign languages across Europe' (LePage & North 2005). These guidelines apply to all languages in Europe. The CEFR breaks a students proficiency into 6 levels; A1, A2, B1, B2,C1and C2. These levels correspond with the traditionally used; beginner (A1), high beginner (A2), up to Advanced (C2) labels. Although the grammar points are different, most of the functions are the same across the languages. I have chosen the CEFR (in this case for English) as it is the only widely agreed upon (albeit tentatively) list of skills and the order in which they should be learned. As far as grammar points and functions go, I should mention that within levels, they are not strictly ordered. For example, there are about 30 grammar points for each level in no particular order. This gives educators some flexibility but it still gives a student the assurance that, once they have achieved mastery of every skill in the level, they are ready to move on to the next set.

My system depends on a common set of skills that can easily be transferred from institution to institution. The CEFR makes all this possible. I should mention here that my system can be easily modified. If the CEFR levels do not match the objectives of a given program, they can easily change the order, add or omit different skills. I reccomend putting any modifications in a different 'course' (I will explain below).  Keeping the uniformity in competency helps when student records are shared. I have taken all of the grammar points and functions from the CEFR and translated them into 'outcomes' in Moodle. Below I will explain how outcomes are used.

Whether or not the definitions for outcomes and competencies are synonymous, the outcomes feature in Moodle serves the purpose of my system well. Like the name implies, once an outcome has been met, we can check it off. In the case of my system, performance can be ranked by how close it is to ideal performance. I have chosen a 5 point Likert scale as I feel it best streamlines the process. My scale is an adaptation of the Interagency Language Roundtable (ILR) (Adam) In my system a 5 = mastery. A student with such a score can perform a given skill with native-like ease. A 4 = acceptable performance. The skill in question is not perfect but it does not require any more work unless the student wants to 'nail it'. A 3 = working performance. The student has studied and is aware of the skill but hasn't mastered it to the point where it can be used without making errors here and there. It basically needs more work. This takes us to 2 which translates to 'limited performance'. A student may or may not be consciously aware of the skill but regardless needs to work on it. Lastly, lets look at 1, minimal performance. The student may not be aware of the skill at all. If there is an awareness, it is at a very elementary level. In Moodle, you will see a hyphen '-' for subjects that have not been taught at all. This can also be considered a 1 and should either be tested or assumed that a student needs this skill.

Description of the System

Let's take an example student to illustrate how this would work. For the purposes of this paper I have chosen 5 famous writers; all from non-English speaking countries. We will be using Gabriel Garcia-Marquez as our example. Like a vast majority of students that will be using this system, he has had some English education in the past. As this system is still in it's developmental stage he will also enter the program without any competency tracking records. At this point, we as educators need to determine his level. Although not a precise match, most of the popular test results (Ielts,TOEIC, TOEFL) have a corresponding CERF level (Baztán). Using results from one of these tests or an in-house test, he can be placed on the CEFR continuum. As an aside, a test is being created for this system as I write this. This should make the system even more self contained and consistent.

In our example, after Mr. Garcia-Marquez's level has been determined, we can start teaching the skills that will take him to the next level. Mr. Garcia-Marquez's level has been determined to be B1 on the CEFR. Before we start his instruction, a more detailed view of his skills within his given level would help us make better curriculum decisions. In Moodle, we have created question banks for each grammar point on the CEFR. As this student placed as a B1, we may want to include questions from the A2 level. Although this student's skills may be at B1 for the most part, there may be gaps here and there. One of the benefits of my proposed system is that,unlike most of the popular tests, it catches these gaps and gives a student useful feedback. Should it be adopted on a large scale, it could also give a more detailed view of a student's abilities to a potential employer or academic program.

Once a students skills are assessed, curriculum decisions can be made about what to teach. This system does not include curriculum content itself, it is up to the respective schools to choose and deliver content. Once the content is taught, competencies can assessed and recorded. As far as grammar is concerned, this program will soon come with question banks and tests that assess grammar. Assessment and lessons can be done online or in class. Should an educator wish to use another instrument, the same scale can be used to keep track of competencies. As for functions, they will need to be assessed by educators in person. The CEFR has a detailed rubric describing how a student should perform a function at each level. Below is an example of how a teacher can rank a student.

The system I am explaining below requires that the LCMS administrator has installed the outcomes. I will explain how to do this later in this paper.

Grading Competencies

Once in a course, in *Settings > Grades* you can access the grade report. Within the grade report, you will see a list of drop-down boxes allowing you to rate a students performance. In the illustration below you see that Mr Garcia-Marquez previously had scored a 3 in his use of 'past modals'. In his last lesson, he demonstrated mastery of the competency so his teacher in this case gave him a 5. With two or three clicks, we have recorded his performance.

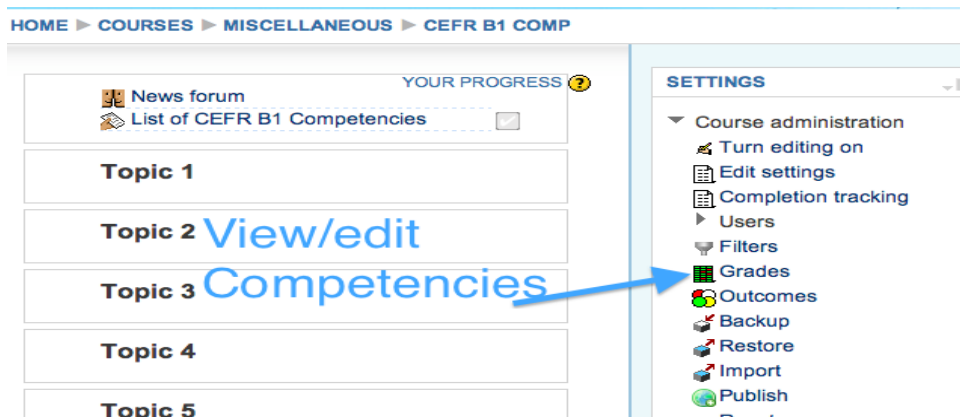

We have seen how this system works with an individual student, lets look at how it works within the context of a class. I have put the list of competencies in a Moodle 'course' as this is the easiest way to organize grades. I should mention that I do not intend for the 'course' to be an actual course but rather to act as a grade-book. This 'course' should be part of a larger syllabus.

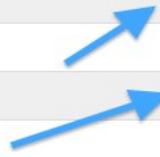

Mr. Garcia-Marquez is in the CERF B1 competencies 'course'. Once again we see Mr. Garcia-Marquez with his writer cohort. If we look in the grader report, we can see what each student has already studied. In the illustration below, if you look to the right, you can see that 3 students have studied the use of 'B1 Should have, might have/etc' . Two students haven't at all as far as we know. A discussion what a program should do in this situation is beyond the scope of this paper but I should point out that, with my system, there is an awareness of the issue. You at least need to be aware of a this kind of issue if you want to find ways to address it. Without any records, I could safely  bet that Mr. Garcia-Marquez and Mr. Rilke would simply repeat the material.

If we look at another competency, in this case 'connecting words expressing cause and effect', we see a different story. As you can see in the illustration below, three students achieved a 3 while the other two haven't studied the topic as far as we know. We can confidently include this topic in the class curriculum.

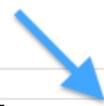
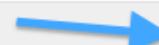
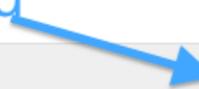

While we are on this page, I would like to point something out. In the lower section of the illustration, you can see one student's name; Ousmane Sembene. As the student's names can be far from their outcomes in the grade-book, the names will appear with a mouse-over. This makes viewing outcomes much easier.

      As I have shown above, a competency tracking system can make curriculum decisions for an institution much more accurate and efficient. For students who want to monitor their own progress, it is also a Godsend. A student needs only to click on the 'Grades' icon in his course to see what he has learned and how well he has mastered it. This gives students more control over their learning. The screen shots below show what Mr. Garcia-Marquez sees as a student when he accesses his grades.

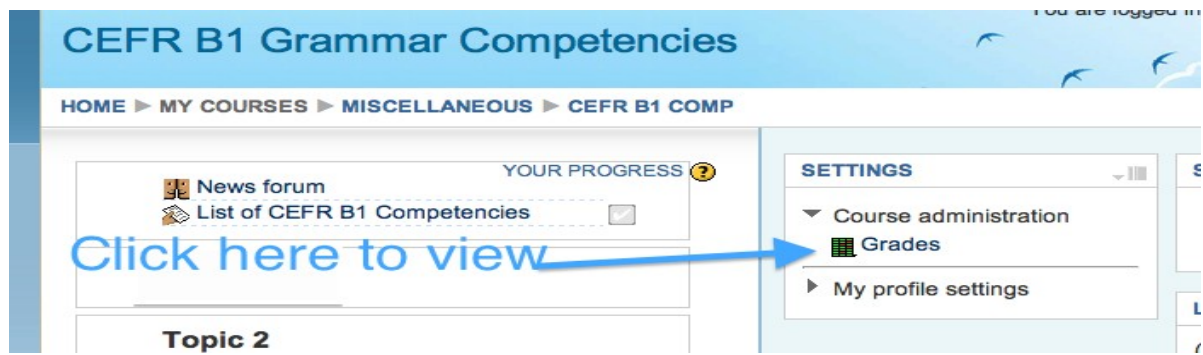

Mr. Garcia-Marquez can see what he has studied and what he needs to study. If you notice in the screen shot above, an educator can also give feedback that can help a student reach his goal. An educator can also export all of this information in spreadsheet format as well. In what follows I will describe how to share this information between institutions.

Exporting and Importing Student Competency Tracking Records

As I mentioned at the beginning of this paper, all of this information can be transferred from one program to the next. In this paper I will discuss such a transfer using Moodle 2.3. It should work from every version onward. I have not tried transferring it to other LCMSs yet but, from what I have seen so far, it seems possible. The data can also be downloaded as a spreadsheet in 2 formats. For the rest of this paper, I will discuss the technical aspects of using the system. To do most of what is mentioned below, you need to have an administrative role or get help from your LCMS administrator.

To use the competency tracking, you first need to enable outcomes in Moodle. You do this by going to *Site Administration > Advanced Features*. At the top of the page, click on the checkbox that says *> Enable Outcomes*. Once enabled, you need to import the competencies included in my program. To do this go to *Settings > Course administration > Outcomes > Outcomes used in course.* Soon, these outcomes will be available online. For now, please send a mail with 'Moodle CEFR' in the title to the e-mail address at the top of this paper. Once you have it, drag and drop file named ALL CEFR outcomes.csv into the file upload area. You can choose whether you would like to import as custom outcomes (for an individual course) or imports as standard outcomes (for site or program wide outcomes). The decision should be based largely on your programs size and its relationship with a larger institution.

You will also need to add Moodle courses to the system. As I mentioned earlier, these will work more as dossiers more than as regular classes. As you will see below, every student's file comes with a 'course' that other students can be added to. The courses themselves should also be put online soon. Please contact me for the courses in the meantime. I have broken up the courses into CEFR levels to make it easier to view a students competencies. We could theoretically include all all levels but this could be much more cumbersome when it comes to viewing competencies. Previous levels can easily be accessed. I have also included a checklist that will will notify educators if all competencies from a previous level have been satisfactorily completed.

With the outcomes and the new courses added to the system, it is now possible to see where a student stands. Although I am trying to find an easier way, for now there is a way, albeit tedious, to export and import a student's CEFR records. First go to the course with the student you want to export. Under *Settings > Course administration >*click *Backup.* Moodle will backup the course with the enrolled students by default so it is only necessary to keep clicking *next.* After you have backed up the class, you will be sent to a page that has the *Course Backup Area.* You now have a file named something like this: backup-Moodle2-course-12-cefr_b1_comp-20130618-0017.mbz. On the left you will see a link that says *restore*. Click that, and you have a copy of the course with all of the students included.

After you have clicked *restore*, you will land on a confirmation page. Click *continue*. This will take you to the *Destination* page. Choose *Restore as a new course,* choose a category to restore the class into and click *continue.* Click *next* through the next few pages. This will result in a new course. The course will have the name 'your course copy 1'. Each following copy will have a consecutive number after the word 'copy', Repeat the above steps for the number of students you would like to export.

My class has five students so I have made 6 copies. I've made an extra copy in case I run into problems. In the copied courses, delete every student but the one you want to make a file for. Change the name of the course to the student's name. In the case of Mr. Garcia-Marquez, the name of the course is 'G Garcia-Marquez CEFR B1 Grammar Competencies' .

Now you have a course with 1 student. Following the steps above, back up this course. After you have backed up the course, you will land on the 'Restore Course' page. Click download as seen in the illustration below. This will give you a copy of the course with Mr. Garcia-Marquez in it. This file can now be shared between programs that use the competency tracking system. As it is a little larger than half a megabyte, it can be easily stored by an institution. It can also be given to the student who can later share it with a future institution when he is ready.

This brings us to the issue of merging students' records together. Let's imagine Mr. Garcia-Marquez has entered a new program. We now need to merge his records with that of the rest of the class. In Moodle 2.3, we need simply to drag and drop his file into the file upload area and click 'restore'. At the bottom of the 'confirm' page click 'next'. This will take you to the 'Destination' page.

In the screen shot above we see have chosen to 'Merge the backup course into the existing course' . We've chosen to put him into the CEFR B1 class where his and the other students in the program can be seen and further tracked. We now have a complete record of all students in our program; what they've done and what they need to do. This gives us salient data that will be very helpful in making curriculum decisions.

This system is just in its infancy. There are lots of improvements I would like to make to help streamline the system. I'd like to see a module that would allow me to import and export an individual student's record without the workaround mentioned above. I'd also like dates automatically added to outcome assessments. Making this process as user friendly as possible is another goal I have for this system. I should mention here that I have avoided using outside plugins or added code for the time being. Using the stock plugins makes it much easier to avoid compatibility issues. I do plan to produce a plugin. I want it to take advantage of capabilities that are built into Moodle as much as possible. Also, if for some reason, the plugin doesn't work in a particular Moodle installation (this happens more that we would like to think!), educators have the workaround to fall back on.

Grammar and functions are just the beginning. I am also planning on adding other aspects of language; vocabulary, listening, writing and anything else that can take advantage of easily taken, shared and viewed student records. The technology is there but I need to find agreed upon criteria for each skill to help with the streamlining of the program. I have chosen to start with grammar and functions as there is agreed upon criteria and it provides a clear view of how the system works.

I am now in the process of piloting it with one institution. This system is particularly suited in this case as the institution involved has small class sizes with students that have varying schedules. The makeup of the group changes from week to week. At the very least, it is helping keep track of where everyone is. I would like to see how it works in others. I am looking forward to feedback that will help me improve on it. I am confident it will be of great benefit to any program that is serious about making sure their students are on the right course. I am also sure that it give students the assurance that they are on the right path.